\documentclass[11pt]{article}

\usepackage{times}
\usepackage[T1]{fontenc}
\usepackage{graphicx}
\usepackage{acl}
\usepackage{microtype}

\usepackage{booktabs}
\usepackage{tabularx}
\usepackage{siunitx}
\usepackage{array}
\usepackage{pifont}
\usepackage{makecell}
\usepackage{amsmath,amssymb}
\usepackage{multirow}
\usepackage{longtable}

\usepackage{listings}
\usepackage{xcolor}
\usepackage{enumitem}
\lstdefinestyle{promptstyle}{
  backgroundcolor=\color{black!5},
  basicstyle=\ttfamily\small,
  breakatwhitespace=false,
  breaklines=true,
  captionpos=b,
  keepspaces=true,
  numbers=left,
  numbersep=5pt,
  xleftmargin=15pt,
  numberstyle=\tiny\color{black!60},
  showspaces=false,
  showstringspaces=false,
  showtabs=false,
  tabsize=2,
  frame=single,
  rulecolor=\color{black!20},
  title=\lstname
}
\usepackage[compact]{titlesec}
\titlespacing{\section}{0pt}{*1}{*1}
\titlespacing{\subsection}{0pt}{*1}{*1}

\usepackage{hyperref}
\usepackage{caption}
\usepackage{float}
\usepackage{tikz}
\usetikzlibrary{positioning, arrows.meta}
\usepackage{subcaption}

\title{MetaGraph: A Large-Scale Meta-Analysis of \\
GenAI in Financial NLP (2022–2025)}

\author{
\textbf{Paolo Pedinotti\thanks{Equal contribution. Paolo Pedinotti contributed to this work during his internship at Bloomberg.}, Peter Baumann, Nathan Jessurun} \\
\textbf{Leslie Barrett, Enrico Santus\footnotemark[1]} \\
Bloomberg \\
\texttt{\{pbaumann25, njessurun, lbarrett4, esantus\}@bloomberg.net}
}

\begin{document}

\maketitle

\begin{abstract}
Financial NLP has evolved rapidly since late 2022, outpacing narrative surveys. We introduce MetaGraph, a methodology for extracting typed knowledge graphs from scientific corpora using ontology-guided LLM extraction to enable structured, large-scale trend analysis. Applied to 681 papers on GenAI in Finance (2022--2025), MetaGraph reveals three phases: early LLM-driven expansion of tasks and datasets, growing emphasis on limitations and risk, and a shift toward modular, system-oriented methods (e.g., retrieval-augmented designs). We release the resulting resource and artifacts to support reproducible meta-analysis and future monitoring of the field.
\end{abstract}

\section{Introduction}

The release of ChatGPT in late 2022 marked a structural shift in NLP, rapidly accelerating the adoption of generative AI (GenAI), particularly large language models (LLMs), in high-stakes domains such as finance. LLMs expanded the scope of Financial NLP beyond traditional supervised pipelines -- long dominated by sentiment analysis and structured extraction -- toward flexible, generative systems capable of zero-shot reasoning, long-document processing, and multimodal inputs.

This rapid evolution has outpaced traditional literature review methodologies. Existing surveys of Financial NLP either predate the widespread use of LLMs or rely on narrative summaries that struggle to capture quantitative trends, structural shifts, and emerging research patterns at scale. As a result, the field lacks a systematic, data-driven view of how tasks, datasets, models, and methods have evolved in response to GenAI.

\begin{figure}[t]
    \centering
    \includegraphics[width=\columnwidth]{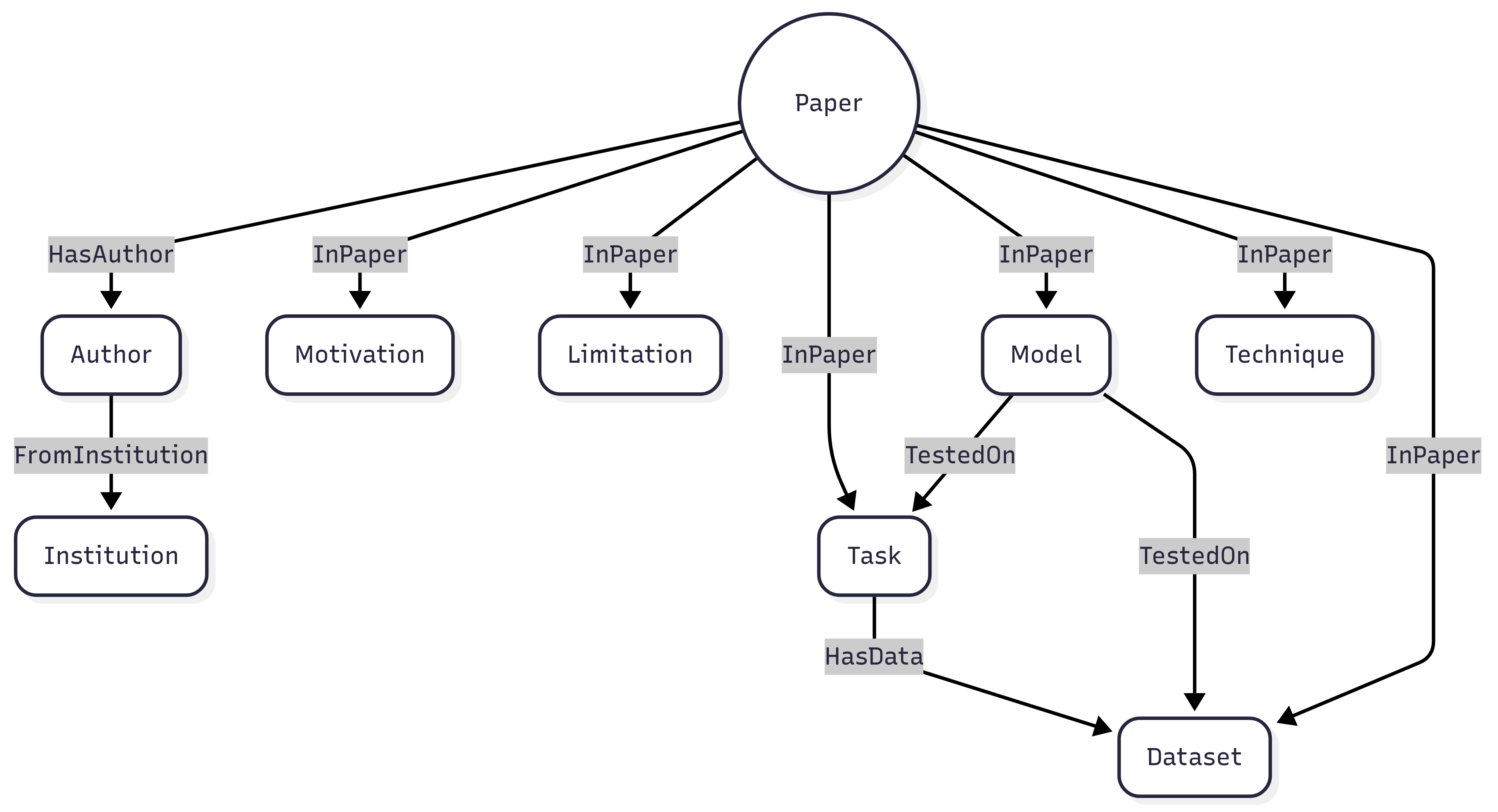}
    \caption{MetaGraph schema: node types and allowed relations (types appear as attributes in the instantiated graph).}
    \label{fig:complete_graph}
\end{figure}

To address this gap, we introduce \textbf{MetaGraph}, a generalizable methodology for extracting structured knowledge graphs from scientific literature using LLMs. MetaGraph combines a manually defined domain ontology with an LLM-based extraction pipeline to transform unstructured papers into a unified, queryable graph capturing research metadata, tasks, datasets, models, techniques, motivations, and limitations. By design, the ontology and prompts are modular, allowing the same framework to be applied beyond Financial NLP. Figure~\ref{fig:complete_graph} shows the high-level structure of the resulting graph.

We apply MetaGraph to a corpus of 681 Financial NLP papers published between 2022 and 2025, producing a structured, queryable representation of the field. This analysis reveals a research landscape undergoing rapid transformation: the rise of financial question answering, the proliferation of datasets enabled by synthetic generation, increasing attention to model limitations and safety, and a gradual shift from model-centric approaches toward integrated systems combining LLMs with retrieval and other auxiliary components.

Our primary contribution is this large-scale, quantitative mapping of GenAI-driven transformation in Financial NLP. MetaGraph serves as the enabling framework that makes such structured, field-level meta-analysis possible. We do not introduce a novel knowledge graph learning algorithm; rather, our contribution lies in operationalizing ontology-guided LLM extraction at scale to produce a reproducible representation of an evolving research domain.

Our contributions are threefold:
(i) \textbf{Field-level analysis}: We provide a data-driven map of Financial NLP’s evolution, tracing systematic shifts in tasks, data sources, model choices, risks, and system design.
(ii) \textbf{Methodology}: We present MetaGraph, a reusable pipeline for ontology-driven extraction and structured meta-analysis of scientific literature.
(iii) \textbf{Open resource}: We release the resulting knowledge graph and artifacts\footnote{The released resource is available at \url{https://zenodo.org/records/16968876}.} 
(provided on acceptance)
to facilitate reproducible meta-analysis and ongoing monitoring of the field.\footnote{58 of the 681 analyzed papers were excluded from redistribution due to CC-BY-NC-ND 4.0 licensing restrictions on arXiv, which prohibit redistribution of derivative content.}

\section{Related Work}

\paragraph{Financial NLP Surveys.} While the survey by \citet{xing2018} centered on traditional tasks such as classification, more recent surveys \citep{li2023, nie2024surveylargelanguagemodels, du2025financial} have focused on the transformative impact LLMs have had on financial applications. Yet, these works rely on a traditional narrative review methodology, qualitatively summarizing the applications in the literature.

Our approach diverges fundamentally by employing a bibliometric and holistic analysis of the field. We uncover structural shifts and data-driven trends by quantitatively mapping the research landscape into a knowledge graph, offering a comprehensive view of the impact of LLMs on its evolution. Table~\ref{tab:survey_comparison} summarizes how MetaGraph complements prior surveys along these dimensions.

\paragraph{LLM-Assisted Knowledge Graph Construction.} \label{app:related_work_llm_knowledge_graph}
\citet{carta2023iterativezeroshotllmprompting} construct domain-specific knowledge graphs through stepwise prompting strategies, while \citet{FunkHJL23} and \citet{10.1007/978-3-031-47240-4_22} use LLMs to learn hierarchical relations among concepts. Our work takes inspiration from GraphRAG \cite{edge2025localglobalgraphrag}, which uses LLMs to extract a knowledge graph and enrich it with information at different levels of granularity.

\begin{table}[t!]
\centering
\small
\setlength{\tabcolsep}{3pt}
\begin{tabular}{@{}l cccc@{}}
\toprule
\textbf{Survey} & \textbf{GenAI} & \textbf{Quantit.} & \textbf{Taxon.} & \textbf{KG} \\
\midrule
Xing et al. (2018)    & \ding{55} & \ding{55} & \ding{55} & \ding{55} \\
Li et al. (2023)     & \ding{51} & \ding{55} & \ding{51} & \ding{55} \\
Nie et al. (2024)     & \ding{51} & \ding{55} & \ding{51} & \ding{55} \\
Du et al. (2025)     & \ding{51} & \ding{55} & \ding{51} & \ding{55} \\
\textbf{Ours (2025)} & \textbf{\ding{51}} & \textbf{\ding{51}} & \textbf{\ding{51}} & \textbf{\ding{51}} \\
\bottomrule
\end{tabular}
\caption{Comparison with Financial NLP surveys.}
\label{tab:survey_comparison}
\end{table}

\section{Methodology}
\label{sec:Methodology}

We introduce \textbf{MetaGraph}, a methodology for automatically constructing knowledge graphs from large scientific corpora to support quantitative analysis. By encoding research entities and their relationships in a structured representation, MetaGraph enables analyses that extend beyond frequency-based statistics and uncover relational patterns that are difficult to detect in unstructured text.

\textbf{MetaGraph} is designed to be \textit{generalizable} and \textit{reusable}, although in this work we apply it specifically to Financial NLP papers.\footnote{Human validation was limited to structured prompt audits on sampled papers and consolidation of taxonomy labels. No large-scale manual correction of extracted entities was performed. The manual validation effort was modest relative to corpus size, and all extraction, normalization, entity resolution, and graph construction steps were fully automated and designed to scale to larger corpora.}

\subsection{Method and Implementation}
MetaGraph builds a typed knowledge graph by (i) defining an ontology; (ii) OCR-ing papers; (iii) extracting schema-constrained records with LLM abstention; (iv) normalizing and resolving entities; (v) inducing taxonomies and enriching metadata; and (vi) instantiating the graph and derived signals (Figure~\ref{fig:complete_graph}).


\subsubsection{Ontology Definition.}
This stage defines the graph structure that determines the analytical scope of MetaGraph. The expressiveness of the ontology directly governs which research patterns can be queried and analyzed.

Our ontology consists of three components: entity types, attributes, and relationships. 

Figure~\ref{fig:complete_graph} summarizes entity types and allowed relations; co-occurrence is inferred via shared \texttt{Paper} links. Graph nodes correspond to the core elements of research papers in the area of NLP, for example Tasks, Models, Methods and Motivations. Narrowing the focus, we investigate topics specifically in Financial NLP allowing a more limited inventory of datasets and tasks.

\subsubsection{Corpus Acquisition}
We compiled 681 Financial NLP papers (November 2022--April 2025) from ACL Anthology and arXiv:

\begin{itemize}[nosep]
    \item \textbf{ACL Anthology:} We queried titles and abstracts of conference papers from 2022--2025 using the official Anthology library, applying an initial filter based on finance-related keywords\footnote{The set of keywords is \textit{financial}, \textit{fintech}, \textit{fraud}, \textit{stock}, \textit{portfolio}, and \textit{finance}} in either the title or abstract. Abstracts were manually screened for relevance to finance as the primary domain.
    
    \item \textbf{arXiv:} Similarly, we utilized the arXiv API to search 2023--2025 preprints in the Computation and Language (cs.CL) and the Quantitative Finance (q-fin) categories using the same set of finance keywords.
\end{itemize}

After aggregating and de-duplicating entries from both sources, manual validation resulted in a final corpus of \textbf{681 papers}. All papers were obtained in PDF format and processed via Mistral OCR\footnote{\url{https://mistral.ai/news/mistral-ocr}} for text extraction. Table~\ref{tab:counts_by_year} reports the number of papers per year.

\begin{table}[t]
\centering
\footnotesize
\setlength{\tabcolsep}{3pt}
\begin{tabularx}{0.60\columnwidth}{>{\raggedright\arraybackslash}X S[table-format=3.0] S[table-format=2.2]}
\toprule
\textbf{Year} & \multicolumn{1}{c}{\textbf{Count}} & \multicolumn{1}{c}{\textbf{Percent (\%)}} \\
\midrule
2022 & 126 & 18.50 \\
2023 & 169 & 24.82 \\
2024 & 273 & 40.09 \\
2025 & 113 & 16.59 \\
\bottomrule
\end{tabularx}
\caption{Number of papers per year in our corpus.}
\label{tab:counts_by_year}
\end{table}

\subsubsection{LLM-based Extraction}
We used Gemini~2.5~Flash\footnote{\url{https://deepmind.google/models/gemini/flash/}} to extract structured information from papers, selecting it for its strong cost-performance trade-off.\footnote{\url{https://lmarena.ai/leaderboard}} The extraction targets standard research entities, including \emph{Tasks}, \emph{Datasets}, \emph{Models}, \emph{Motivations}, \emph{Techniques}, \emph{Limitations}, and \emph{Author} and \emph{Institution} metadata.

We refined prompts via small-sample audits targeting omissions, over-generation, and schema violations; updates emphasized task definitions, abstention, and output constraints (Appendix~\ref{apx:prompts}).
Based on the observed patterns, we updated the prompt instructions (e.g., clarifying task definitions, strengthening abstention rules, and constraining output formats). Refinement stopped once the audit showed no recurring systematic errors. The full set of prompts is provided in Appendix~\ref{apx:prompts}.

\paragraph{Multi-instance representation.}
Each \texttt{Paper} links to multiple \texttt{Task}/\texttt{Dataset}/\texttt{Model} nodes (many-to-many). Figure~\ref{fig:multi_instance_example} illustrates this structure.

\begin{figure}[t]
\centering
\resizebox{1\columnwidth}{!}{
\begin{tikzpicture}[
    >=Latex,
    node distance=1.0cm and 2.4cm,
    edge/.style={-Latex, thick},
    paper/.style={draw, rounded corners, align=center, font=\footnotesize},
    dataset/.style={draw, rounded corners, align=center, font=\footnotesize, fill=blue!10},
    model/.style={draw, rounded corners, align=center, font=\footnotesize, fill=green!12},
    lbl/.style={font=\scriptsize, fill=white, inner sep=1pt}
]

\node[paper] (paper) {Paper $P$};
\node[dataset, right=of paper] (data) {FinQA};
\node[model, right=of data, yshift=0.95cm] (m1) {GPT-4};
\node[model, right=of data, yshift=-0.95cm] (m2) {FinBERT};

\draw[edge] (paper) -- node[lbl, above] {UsesDataset} (data);

\draw[edge, bend left=18] (paper) to node[lbl, above] {EvaluatesModel} (m1);
\draw[edge, bend right=18] (paper) to node[lbl, below] {EvaluatesModel} (m2);

\draw[edge, bend left=12] (m1) to node[lbl, right] {TestedOn} (data);
\draw[edge, bend right=12] (m2) to node[lbl, right] {} (data);

\end{tikzpicture}
}
\caption{Example of multi-instance representation in the knowledge graph: a paper evaluates multiple models on a dataset (datasets in blue, models in green).}
\label{fig:multi_instance_example}
\end{figure}

\paragraph{Validation.} 
Validation on 12 gold papers found only minor omissions (two tasks, one model) and no hallucinations.

\subsubsection{Entity Resolution}
\label{sec:entity_resolution}

We resolved surface-form inconsistencies using normalization followed by embedding-based clustering. Mentions were first lowercased and stripped of punctuation and formatting artifacts. For each entity type independently (e.g., \texttt{Dataset}, \texttt{Model}), normalized mentions were embedded using OpenAI’s \texttt{text-embedding-small}. Pairwise cosine similarity was computed, and mentions were merged if cosine similarity was $\geq 0.93$, using greedy agglomeration. 

The threshold $\tau=0.93$ was chosen conservatively to avoid merging semantically related but distinct entities. For example, \textit{Finqa} $\rightarrow$ \textit{FinQA}, \textit{FIQA-SA} variants $\rightarrow$ \textit{FIQA-SA}, while \textit{FinQA} and \textit{ConvFinQA} remained distinct.


\subsubsection{Taxonomy Induction}
Selected entity types were organized into taxonomies using a zero-shot LLM-based categorization procedure. We processed batches of maximum 100 entities at a time. For each batch, the LLM was prompted to propose recurring category types and assign entities accordingly. We manually consolidated categories by merging near-duplicates, removing singletons, and standardizing naming. This final step affected only taxonomy naming accuracy; entity extraction remained fully automated.

\paragraph{Institution type labeling.}
Institutions were labeled as \textit{academic}, \textit{industry}, or \textit{mixed} using rule-based heuristics applied to affiliation strings and Semantic Scholar metadata\footnote{\url{https://api.semanticscholar.org/}}. Keywords such as \textit{University} and \textit{College} triggered the \textit{academic} label, while terms such as \textit{Inc.}, \textit{Ltd.}, etc., triggered \textit{industry}. Papers with affiliations spanning categories were labeled \textit{mixed}. The procedure was fully deterministic and did not rely on LLM inference.



\subsubsection{Relevance Scoring}
To highlight emerging trends, we define a paper-level \textit{relevance score} used to prioritize analyses toward more influential work. The score combines three factors: (i) \textbf{institutional centrality}, computed as the PageRank of affiliated institutions in the co-authorship graph; (ii) \textbf{productivity}, measured as the number of papers published by the institution; and (iii) \textbf{citation normalization}, defined as paper citations normalized by the average citation count of the publication year.

\section{Findings and Insights}

\begin{figure}
  \centering
  \includegraphics[width=1\columnwidth]{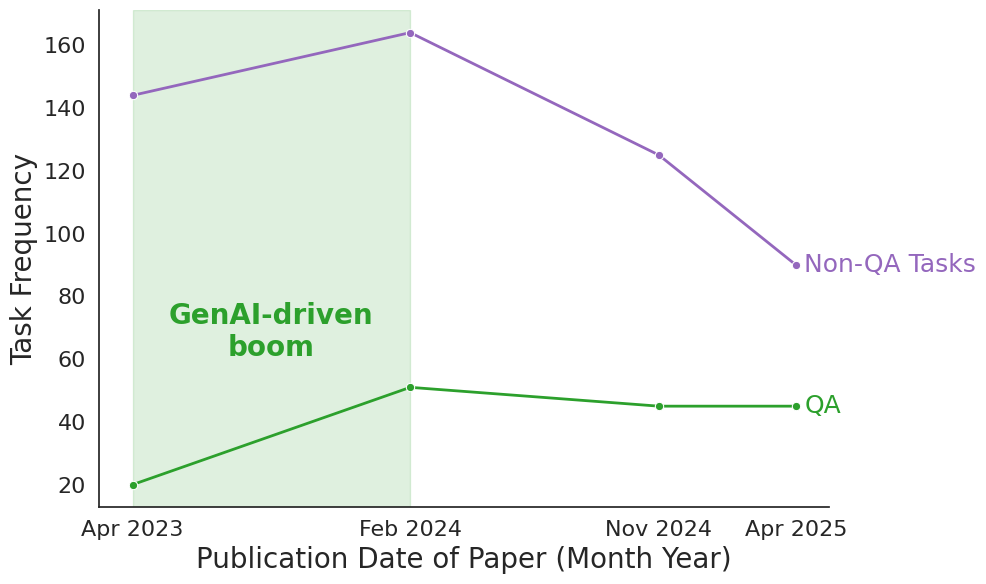}
  \caption{Increasing focus on financial QA. Task frequency here is the number of papers with an instance of the task category.}
  \label{fig:treated_task_evolution}
\end{figure}

In this section, we demonstrate the types of analyses enabled by MetaGraph, focusing on how the release of ChatGPT in late 2022 marked a turning point for Financial NLP. For comparability, we partition the corpus into three chronological subsets of approximately equal size: T1 (January~2022--August~2023), T2 (September~2023--July~2024), and T3 (August~2024--April~2025). All temporal analyses in this section use these fixed partitions. The main findings are:
\begin{itemize}[nosep]
    \item Task emphasis shifts toward QA variants (Figure~\ref{fig:treated_task_evolution}; Table~\ref{tab:evolution_task_selected}).
    \item The dataset landscape fragments and data sources diversify (Table~\ref{tab:dataset_periods}; Table~\ref{tab:evolution_dsets_sources_and_signals}).
    \item Reported limitations shift from data scarcity toward model- and safety-related concerns (Figure~\ref{fig:evolution_limitations}; Table~\ref{tab:evolution_safety_concerns}).
    \item Methods evolve from prompting-centric approaches toward system-level designs (Figure~\ref{fig:techniques_evolution}; Table~\ref{tab:rag_evolution}).
\end{itemize}

More details and insights are provided in Appendices~\ref{apx:geography}, ~\ref{apx:prompts}, ~\ref{apx:analysis}.

\subsection{Release of ChatGPT}
Before the release of ChatGPT, the field focused mainly on sentiment analysis, information extraction, and stock prediction. In the period immediately following the release of ChatGPT (November 2022--February 2024), these tasks still constituted 90\% of published work, and the most widely used datasets (Table~\ref{tab:dataset_periods_early}) reflected this focus.

As usage of LLMs matured (February 2024--April 2025, Table~\ref{tab:dataset_periods_late}), the landscape evolved unlocking new applications and attention toward more complex and reasoning-intensive tasks. \textbf{Financial QA} has become the leading focus, rising from 10\% to 33\% of tasks by 2025, while traditional tasks have steadily declined, as shown in Table~\ref{tab:evolution_task_selected} for representative tasks (temporal distribution of the full taxonomy in Appendix Table~\ref{tab:evolution_task_variants}).

LLMs have transformed \textbf{the way} researchers approach financial problems. This shift moves from narrow, task-specific pipelines to flexible, generative systems that bridge previously isolated tasks. Between April 2023 and February 2024, the average number of tasks per paper rose from 1.36 to 1.9. Traditional tasks such as sentiment analysis and information extraction are now often used as intermediate steps in broader systems, such as RAG and financial agents.

\begin{figure}
  \centering
  \includegraphics[width=1\columnwidth]{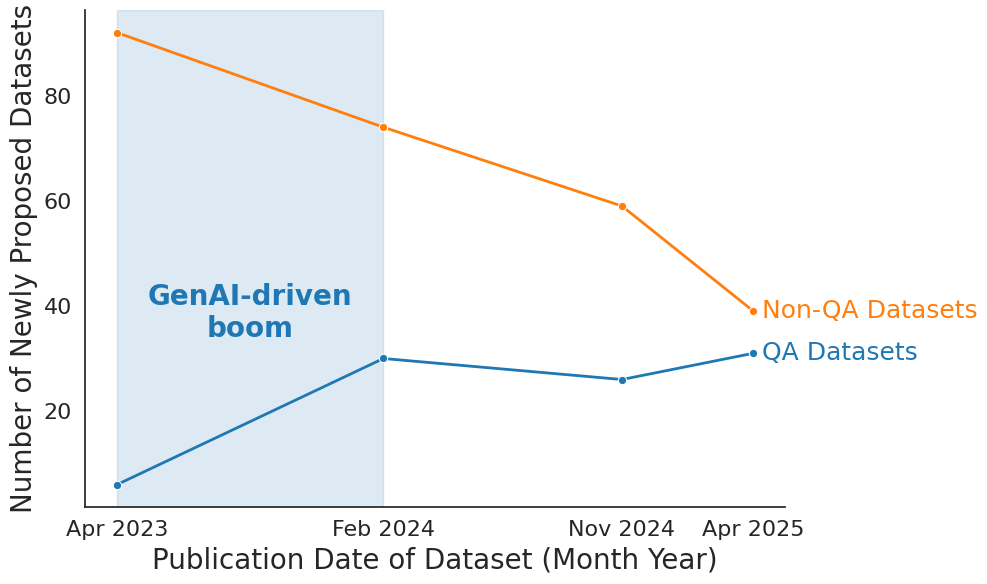}
  \caption{New datasets by period.}
  \label{fig:created_data_evolution}
\end{figure}

\paragraph{Data Sources and Datasets.}
Datasets evolved as well. On the one hand, QA benchmarks now lead the field, overtaking traditional datasets (Table~\ref{tab:dataset_periods}). On the other hand, we witnessed an expansion and diversification of the data sources used to generate QA datasets. Recent papers increasingly mention multimodal and structured inputs -- such as tables, charts, audio, and analyst commentaries -- alongside core sources such as news and company reports (Table~\ref{tab:evolution_dsets_sources_and_signals}). This expansion has been supported by synthetic data generation, which reduced the need for expert annotations. The share of synthetic or human-in-the-loop datasets nearly tripled as LLMs became data generators, from 5\% in April 2023 to almost 15\% by November 2024 \citep{krumdick-etal-2024-bizbench, guo-yang-2024-econnli, liu2025evaluatingaligninghumaneconomic, li2023cfgptchinesefinancialassistant}. Data trends for tasks are plotted in Figure~\ref{fig:created_data_evolution}. Most new datasets target QA tasks, while the development of datasets for other tasks, such as sentiment analysis, has slowed.

\subsection{A Growing Awareness}
\begin{figure}
  \centering
  \includegraphics[width=1\columnwidth]{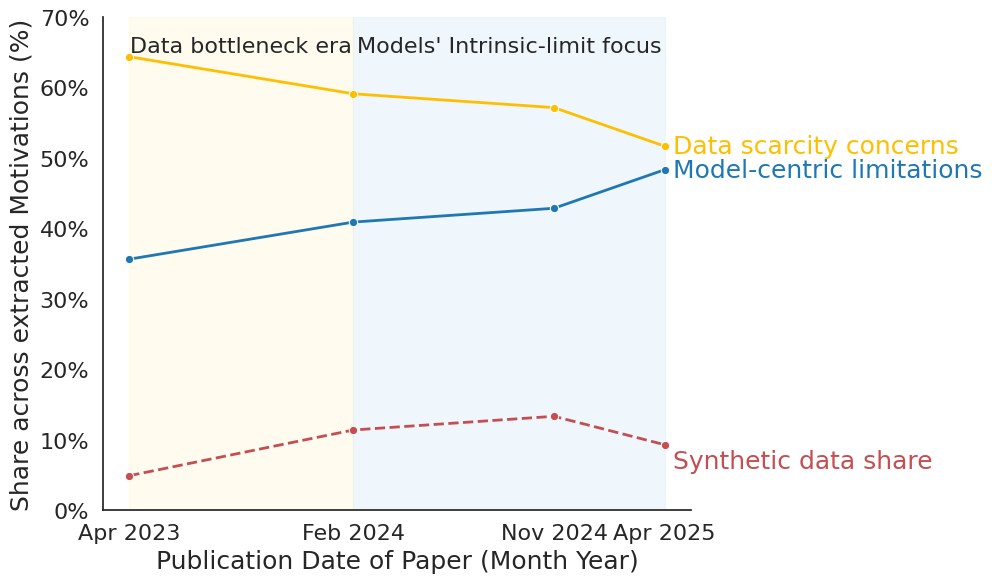}
  \caption{Reported limitations by period. Synthetic data share complements data scarcity concerns.}
  \label{fig:evolution_limitations}
\end{figure}

LLMs have lowered key barriers to both adoption and data processing. On the one hand, they remove data format constraints -- enabling the processing of unstructured data. On the other hand, they support synthetic data generation, helping mitigate challenges such as cost, scarcity, and domain bias (Table~\ref{tab:evolution_limitations}). We show how limitations have changed over time in Figure~\ref{fig:evolution_limitations}. As data constraints eased, research attention increasingly shifted toward model-level challenges -- particularly reasoning, interpretability, efficiency, and safety. We observed growing concerns around bias, privacy risks, and potential misuse (Table~\ref{tab:evolution_safety_concerns}).

This shift toward critical reflection is evident in the evolution of research motivations, which increasingly convey a more cautious stance toward LLMs. By 2024, critical themes such as robustness, efficiency, reasoning, and RAG appeared in nearly 18\% of papers -- twice the share observed in early 2023 (see Table~\ref{tab:evolution_motivation}). This marks a shift from earlier studies, which primarily focused on leveraging LLMs through zero-shot learning and fine-tuning.

This marks a move from standalone \textit{LLMs} to \textit{system-oriented design}. Prompting strategies have evolved as well (Table~\ref{tab:evolution_prompt_engineering}): the progression from in-context learning to augmented methods such as chain-of-thought, retrieval-based prompts, and self-criticism reflects a move away from relying solely on the model’s few-shot capabilities toward more deliberate prompt enrichment aimed at reducing errors.

\begin{table}
\centering
\footnotesize
\setlength{\tabcolsep}{3pt}
\begin{tabularx}{\columnwidth}{X S[table-format=2.2] S[table-format=2.2] S[table-format=2.2]}
\toprule
\textbf{Task} & \textbf{T1} & \textbf{T2} & \textbf{T3} \\
\midrule
Financial Sentiment Analysis & 15.67 & 15.77 & 9.78 \\
NER & 7.21 & 8.67 & 5.32 \\
Stock Price Change Prediction & 12.19 & 7.63 & 8.20 \\
Retrieval Enhanced QA & 5.47 & 7.80 & 12.81 \\
Numerical QA & 5.47 & 7.45 & 9.50 \\
Long Document QA & 3.23 & 4.51 & 7.34 \\
Financial Consulting QA & 2.24 & 4.16 & 6.04 \\
Claim Verification & 1.24 & 1.21 & 3.31 \\
\bottomrule
\end{tabularx}
\caption{Distribution (in \% of papers) of selected Financial NLP tasks across time periods (T1: Jan 2022--Aug 2023, T2: Sep 2023--Jul 2024, T3: Aug 2024--Apr 2025).}
\label{tab:evolution_task_selected}
\end{table}

\begin{table*}
\centering
\begin{subtable}[t]{0.48\textwidth}
\centering
\small
\begin{tabular}{|l|l|c|}
\hline
\textbf{Dataset} & \textbf{Task} & \textbf{Freq.} \\
\hline
FPB \cite{https://doi.org/10.1002/asi.23062} & SA & 29 \\
FinQA \cite{chen-etal-2021-finqa}             & QA & 19 \\
FIQA-SA  \cite{10.1145/3184558.3192301}       & SA & 15 \\
ConvFinQA \cite{chen-etal-2022-convfinqa}     & QA & 13 \\
RefInd \cite{10.1145/3539618.3591911}         & RE & 7  \\
\hline
\end{tabular}
\caption{November 2022 -- February 2024}
\label{tab:dataset_periods_early}
\end{subtable}
\hfill
\begin{subtable}[t]{0.48\textwidth}
\centering
\small
\begin{tabular}{|l|l|c|}
\hline
\textbf{Dataset} & \textbf{Task} & \textbf{Freq.} \\
\hline
ConvFinQA \cite{chen-etal-2022-convfinqa}      & QA & 13 \\
FPB \cite{https://doi.org/10.1002/asi.23062}   & SA & 13 \\
FinQA   \cite{chen-etal-2021-finqa}            & QA & 12 \\
FIQA-SA \cite{10.1145/3184558.3192301}         & SA & 7  \\
FinanceBench \cite{islam2023financebenchnewbenchmarkfinancial} & QA & 7 \\
\hline
\end{tabular}
\caption{February 2024 -- April 2025}
\label{tab:dataset_periods_late}
\end{subtable}
\caption{Top datasets in Financial NLP by usage across two time periods. (QA: Question Answering, SA: Sentiment Analysis, RE: Relation Extraction).} 
\label{tab:dataset_periods}
\end{table*}

\begin{figure}
    \centering
    \includegraphics[width=1\columnwidth]{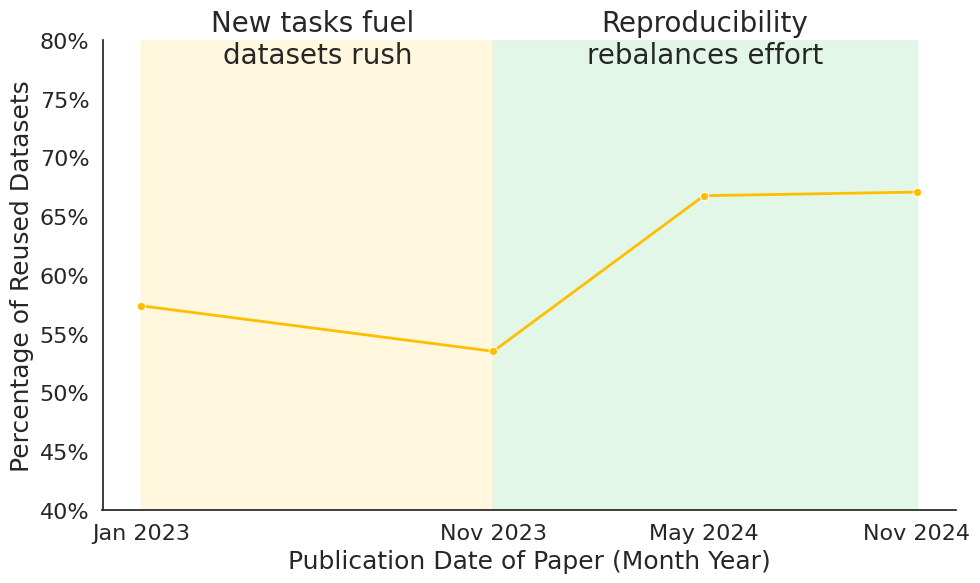}
    \caption{Temporal evolution of the proportion of datasets with references to prior literature.}
    \label{fig:created_data_evolution_2}
\end{figure}

\begin{figure}
  \centering
  \includegraphics[width=1\columnwidth]{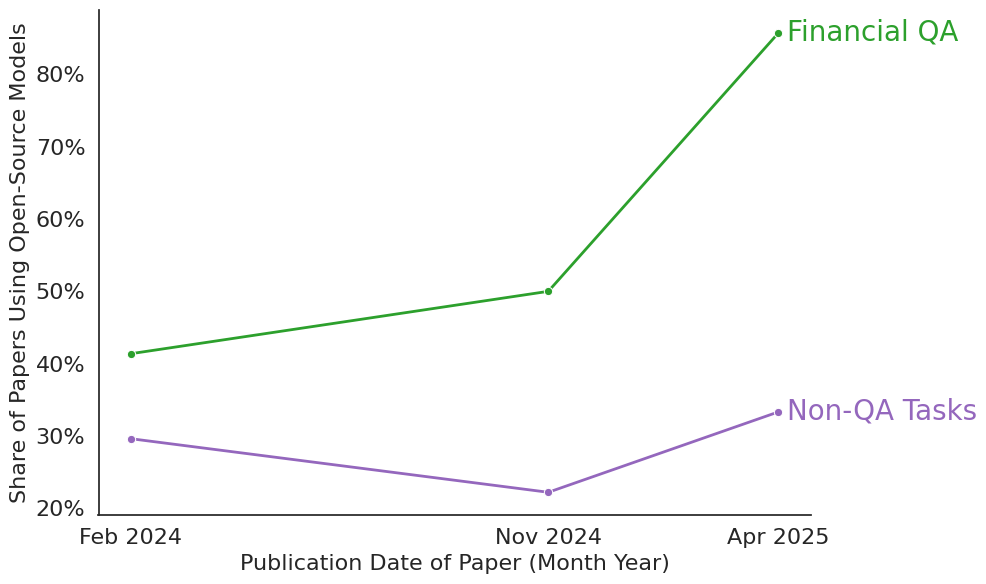}
  \caption{Share of papers using open-source models by task and timeframe.}
  \label{fig:open_models_evolution}
\end{figure}

\begin{figure}
  \centering
  \includegraphics[width=1\columnwidth]{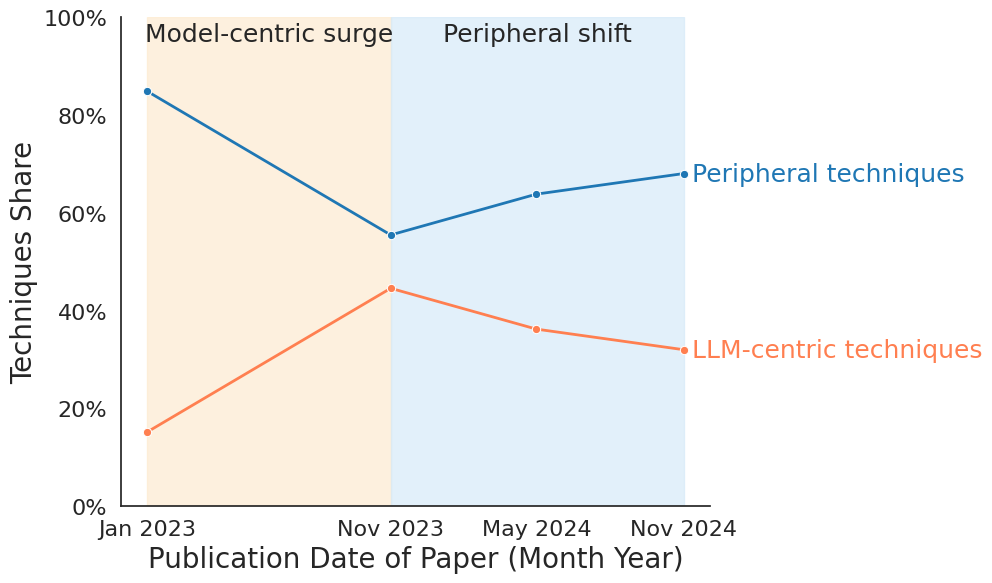}
  \caption{Technique evolution over time.}
  \label{fig:techniques_evolution}
\end{figure}

\subsection{From Models to Systems}

In the wake of GenAI’s rise, researchers initially focused on adapting general-purpose LLMs to Financial NLP tasks through prompt engineering -- especially zero-shot and in-context learning -- which quickly gained momentum across applications. This was often complemented by post-training methods such as instruction tuning to further specialize models for the financial domain (Table~\ref{tab:evolution_prompt_engineering}).

Researchers began to move beyond model-centric approaches as the limitations of reasoning, safety, interpretability, and scalability became more apparent. Over time, these techniques were increasingly complemented by \emph{system-level innovations} that integrate LLMs into broader frameworks (Figure~\ref{fig:techniques_evolution}).

The most prominent of these is RAG \cite[among others]{10.1145/3604237.3626866, xue2024weaverbirdempoweringfinancialdecisionmaking, li-etal-2024-alphafin, yepes2024financialreportchunkingeffective, chen-etal-2024-knowledge, zhao-etal-2024-optimizing}, which has become a cornerstone of the field. Examining the evolution of RAG (Table~\ref{tab:rag_evolution}), we find it mirrors the dataset trend: the spectrum of source types and data formats has widened, knowledge bases have grown, and the size of retrieved context has expanded from single sentences to large document chunks.

\subsection{Towards Maturity}

As the field matured, researchers began prioritizing shared resources over creating new \textbf{datasets}, increasingly relying on established, literature-backed benchmarks (Figure~\ref{fig:created_data_evolution_2}), with growing coverage across tasks.
A similar trend emerged on the modeling side, as the community increasingly turned to open-source models valued for their transparency, controllability, and adaptability, alongside a shift from rapid expansion toward critical evaluation (Figure~\ref{fig:open_models_evolution}).

\begin{figure}[h]
  \centering
  \includegraphics[width=\columnwidth]{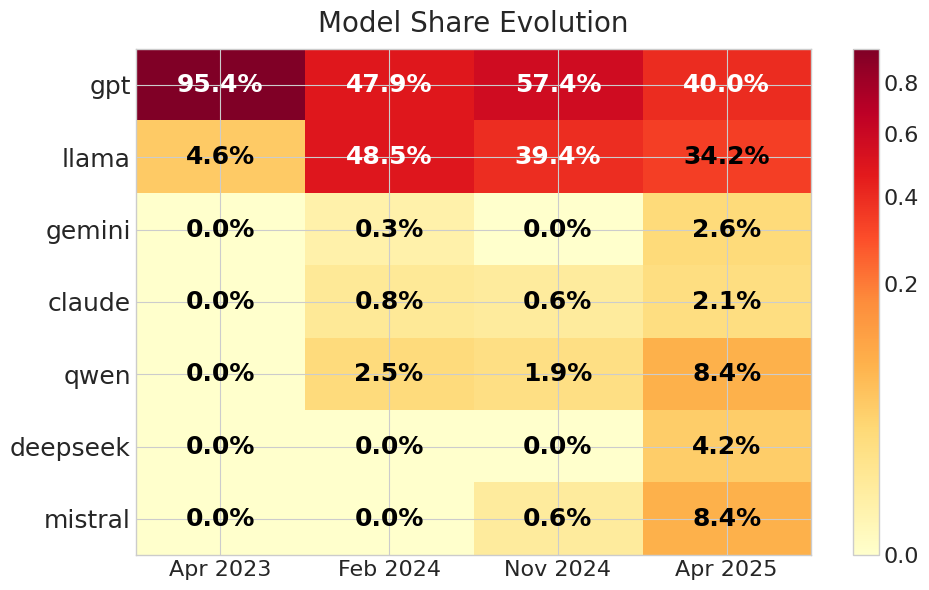}
  \caption{LLMs usage distribution over time.}
  \label{fig:models_evolution}
\end{figure}

\begin{figure}[h]
  \centering
  \includegraphics[width=\columnwidth]{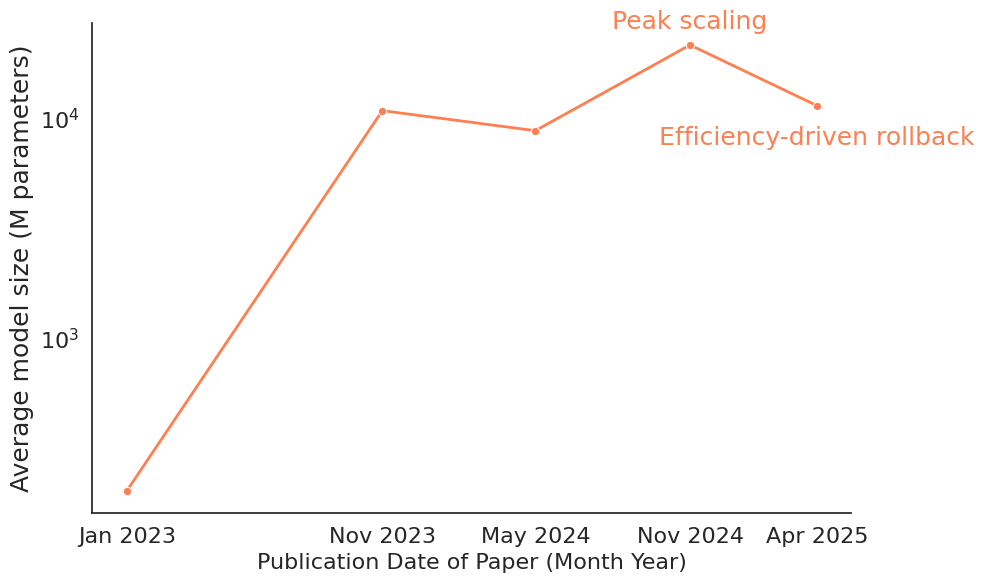}
  \caption{Sizes of open-source LLMs over time.}
  \label{fig:scaling}
\end{figure}

Figure~\ref{fig:models_evolution} illustrates three key phases: the early dominance of GPT models, the emergence of LLaMA \cite{touvron2023llamaopenefficientfoundation}, and the current diversification toward a mix of open models -- such as Qwen \cite{bai2023qwentechnicalreport} and DeepSeek \cite{deepseekai2025deepseekr1incentivizingreasoningcapability} -- and proprietary ones. Figure~\ref{fig:scaling} shows how model sizes have also changed over time. The field is revisiting cost-performance trade-offs, driven by the financial and computational cost of large models. This shift is reflected in a recent inflection in model size trends.

\begin{table}
\centering
\footnotesize
\setlength{\tabcolsep}{3pt}
\begin{tabularx}{\columnwidth}{
  >{\raggedright\arraybackslash}X
  S[table-format=2.2]
  S[table-format=2.2]
  S[table-format=2.2]
}
\toprule
\textbf{Data} & \textbf{T1} & \textbf{T2} & \textbf{T3} \\
\midrule
\textbf{Sources} & \textbf{(\%)} & \textbf{(\%)} & \textbf{(\%)} \\
\cmidrule(l){1-4}
News                             & 27.48 & 29.14 & 25.35 \\
Social Media / Forums            & 21.85 & 14.20 & 14.43 \\
Company Reports                  & 28.15 & 27.37 & 28.99 \\
Company Fundamentals \& Indicators & 11.04 & 15.09 & 15.13 \\
Earnings Calls                   & 5.18  & 7.10  & 6.58  \\
Analyst Reports                  & 4.73  & 3.25  & 3.92  \\
University Textbooks             & 1.13  & 0.89  & 2.38  \\
Financial Analyst Exams          & 0.45  & 2.96  & 3.22  \\
\midrule
\multicolumn{4}{l}{\textbf{Signals}} \\
\cmidrule(l){1-4}
Text                             & 80.42 & 73.72 & 70.44 \\
Tables                           & 16.08 & 19.87 & 20.13 \\
Image                            & 1.40  & 2.56  & 5.03  \\
Audio                            & 0.70  & 1.92  & 1.89  \\
Other                            & 2.10  & 1.92  & 2.52  \\
\bottomrule
\end{tabularx}
\caption{Distribution of data sources and signal types across time periods (T1: Jan 2022--Aug 2023, T2: Sep 2023--Jul 2024, T3: Aug 2024--Apr 2025).}
\label{tab:evolution_dsets_sources_and_signals}
\end{table}

\begin{table}
\centering
\small
\setlength{\tabcolsep}{3pt}
\begin{tabularx}{\columnwidth}{
  >{\raggedright\arraybackslash}X
  S[table-format=3.0]
  S[table-format=3.0]
  S[table-format=3.0]
}
\toprule
\textbf{Risk / Limitation} & {\textbf{T1}} & {\textbf{T2}} & {\textbf{T3}} \\
\midrule
\multicolumn{4}{l}{\textbf{Dominant Issues}} \\
\cmidrule(r){1-4}
Misleading Predictions due to Outdated Data & 60 & 100 & 119 \\
Misleading Predictions due to Inaccuracies & 50 & 66 & 75 \\
Lack of Transparency/Explainability & 27 & 66 & 94 \\
Inability to Generalize Across Financial Tasks & 27 & 58 & 64 \\
\midrule
\multicolumn{4}{l}{\textbf{Growing Concerns}} \\
\cmidrule(r){1-4}
Biases Towards Stocks/Trends/Products & 12 & 30 & 43 \\
Breaches of Sensitive Data & 4 & 16 & 25 \\
Susceptibility to Attacks and Misuse & 10 & 12 & 25 \\
Misinterpretation of Regulatory Text & 1 & 7 & 21 \\
Gender or Demographic Bias & 6 & 21 & 17 \\
Generation of Fraudulent Content & 3 & 7 & 15 \\
\midrule
\multicolumn{4}{l}{\textbf{Stable Low-Level Issues}} \\
\cmidrule(r){1-4}
Sensitivity to Data/Market Shifts & 25 & 36 & 33 \\
Inability to Detect Misinformation & 5 & 7 & 8 \\
Susceptibility to Corpus Poisoning & 0 & 9 & 3 \\
\bottomrule
\end{tabularx}
\caption{Classification of LLM risks in finance based on their frequency over time periods (T1: Jan 22--Aug 23, T2: Sep 23--Jul 24, T3: Aug 24--Apr 25).}
\label{tab:evolution_safety_concerns}
\end{table}

\begin{table}
\centering
\small
\setlength{\tabcolsep}{3pt}
\begin{tabularx}{\columnwidth}{
  >{\raggedright\arraybackslash}X
  S[table-format=2.2]
  S[table-format=2.2]
  S[table-format=2.2]
}
\toprule
\textbf{RAG Configuration} & {\textbf{T1}} & {\textbf{T2}} & {\textbf{T3}} \\
\midrule
\multicolumn{4}{l}{\textbf{Data Source Size}} \\
\cmidrule(r){1-4}
Small & 38.24 & 33.33 & 34.15 \\
Medium & 20.59 & 17.54 & 14.63 \\
Large & 41.18 & 49.12 & 51.22 \\
\midrule
\multicolumn{4}{l}{\textbf{Data Source Type}} \\
\cmidrule(r){1-4}
Text & 66.67 & 64.71 & 60.51 \\
Table & 18.33 & 21.57 & 20.38 \\
Database & 5.00 & 7.84 & 10.83 \\
Graph & 8.33 & 2.94 & 5.10 \\
Image & 1.67 & 2.94 & 2.55 \\
Other & 0.00 & 0.00 & 0.64 \\
\midrule
\multicolumn{4}{l}{\textbf{Retrieved Text Granularity}} \\
\cmidrule(r){1-4}
Chunk & 84.85 & 87.72 & 95.18 \\
Sentence & 15.15 & 12.28 & 4.82 \\
\bottomrule
\end{tabularx}
\caption{Distribution of RAG configurations across time periods.} 
\label{tab:rag_evolution}
\end{table}

\begin{table}
\centering
\footnotesize
\setlength{\tabcolsep}{3pt}
\begin{tabularx}{\columnwidth}{
  >{\raggedright\arraybackslash}X
  S[table-format=2.2]
  S[table-format=2.2]
  S[table-format=2.2]
}
\toprule
\textbf{Data} & \textbf{T1} & \textbf{T2} & \textbf{T3} \\
\midrule
\multicolumn{4}{l}{\textbf{Data-related Limitations}} \\
Costly Human Judging                    & 3.20 & 3.22 & 2.83 \\
Insufficient Data Scale/Coverage        & 12.45 & 13.00 & 10.59 \\
Skewed/Imbalanced Classes               & 3.08 & 2.29 & 1.47 \\
Domain/Language Bias                    & 12.03 & 11.69 & 9.59 \\
\midrule
\multicolumn{4}{l}{\textbf{LLM Limitations}} \\
Interpretability Gaps                   & 1.63 & 2.29 & 2.04 \\
Weak Reasoning                          & 4.11 & 4.59 & 5.24 \\
Cost \& Environmental Footprint         & 2.66 & 2.79 & 3.46 \\
Hallucination \& Bias                   & 2.18 & 2.95 & 3.72 \\
Prompt Sensitivity                      & 1.75 & 2.84 & 2.78 \\
Latency / Scalability                   & 2.06 & 2.18 & 2.57 \\
Synthetic Data / Label Issues           & 7.38 & 4.86 & 5.14 \\
Capacity Constraints                    & 9.07 & 9.07 & 9.70 \\
\midrule
Gaps: Lab vs.\ Live                     & 9.43 & 8.68 & 10.27 \\
Other (Appendix~\ref{app:risks_and_limitations})                    & 28.81 & 29.55 & 30.06 \\
\bottomrule
\end{tabularx}
\caption{Distribution of reported limitations across time periods.} 
\label{tab:evolution_limitations}
\end{table}

\begin{table}
\centering
\footnotesize
\setlength{\tabcolsep}{3pt}
\begin{tabularx}{\columnwidth}{
  >{\raggedright\arraybackslash}X
  S[table-format=2.2]
  S[table-format=2.2]
  S[table-format=2.2]
}
\toprule
\textbf{Data} & \textbf{T1} & \textbf{T2} & \textbf{T3} \\
\midrule
\multicolumn{4}{l}{\textbf{Data}} \\
Data Scarcity \& Annotation Cost     & 32.25 & 28.82 & 23.00 \\
Other                                & 37.12 & 35.81 & 35.48 \\
\midrule
\multicolumn{4}{l}{\textbf{Exploiting LLMs}} \\
Zero/Few-Shot Evaluation             & 4.41  & 4.37  & 2.53 \\
Domain-Specific LLM Training         & 10.44 & 8.95  & 11.89 \\
\midrule
\multicolumn{4}{l}{\textbf{Solving LLM Limitations}} \\
Quantitative Reasoning Gaps          & 5.10  & 5.02  & 6.43 \\
Interpretability \& Explainability   & 3.25  & 3.71  & 5.07 \\
Efficiency Constraints               & 3.02  & 5.24  & 5.65 \\
Safety, Robustness, \& Fairness      & 2.78  & 5.90  & 7.21 \\
RAG \& Retrieval Bottlenecks         & 1.62  & 2.18  & 2.73 \\
\bottomrule
\end{tabularx}
\caption{Distribution of future research directions across time periods.} 
\label{tab:evolution_motivation}
\end{table}

\begin{table}
\centering
\footnotesize
\setlength{\tabcolsep}{3pt}
\begin{tabularx}{\columnwidth}{
  >{\raggedright\arraybackslash}X
  S[table-format=3.0]
  S[table-format=3.0]
  S[table-format=3.0]
}
\toprule
\textbf{Technique} & \textbf{T1} & \textbf{T2} & \textbf{T3} \\
\midrule
\multicolumn{4}{l}{\textbf{Core Prompting}} \\
Zero-Shot & 69 & 126 & 162 \\
Few-Shot & 20 & 98 & 74 \\
Chain-of-Thought & 13 & 61 & 94 \\
\midrule
\multicolumn{4}{l}{\textbf{Augmentation Strategies}} \\
RAG (Retrieval-Augmented) & 39 & 65 & 92 \\
Decomposition & 23 & 52 & 58 \\
Self-Criticism & 4 & 13 & 30 \\
\midrule
\multicolumn{4}{l}{\textbf{High-Level Methods}} \\
Ensembling & 17 & 14 & 13 \\
Agents & 9 & 25 & 40 \\
\bottomrule
\end{tabularx}
\caption{Frequency counts of prompt engineering techniques across time periods.}
\label{tab:evolution_prompt_engineering}
\end{table}

\begin{figure}
  \centering
  \small
  \includegraphics[width=1\columnwidth]{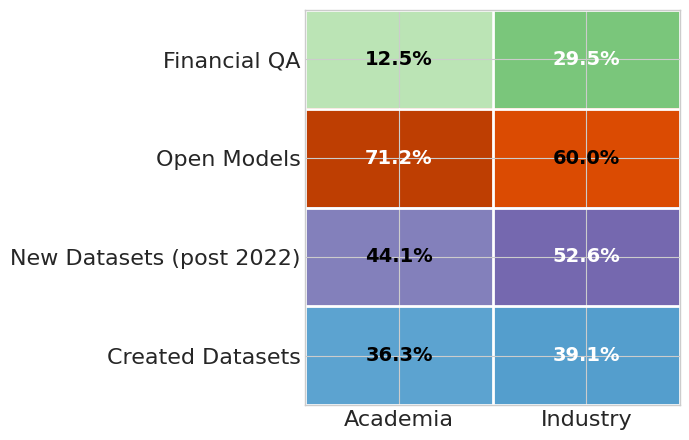}
  \caption{Academia vs.\ Industry: proportions of instances of tasks, models, and datasets.}
  \label{fig:academia_vs_industry}
\end{figure}

\textbf{One Revolution, Two Speeds.}
GenAI reshaped industry and academia at different paces (Figure~\ref{fig:academia_vs_industry}). We took all the instances of tasks, models, and datasets in our corpus, and computed the relative proportion of financial QA instances, open-model instances, new datasets (datasets created after 2022), and created datasets (datasets created by the same authors who use them). Industry moved faster -- dominating financial QA and driving dataset innovation to stay competitive. Academia responded more cautiously, focusing on established tasks and open-source models, with a stronger emphasis on transparency and reproducibility. This is likely due to academia’s structural constraints, which prioritize transparency, reproducibility, and the use of publicly available data and models -- factors that inherently slow down the adoption of cutting-edge approaches. In contrast, industry has largely traded off transparency in favor of rapid experimentation, leveraging proprietary data and closed-source LLMs to push forward advanced use cases such as financial QA.

\subsection{Looking Ahead}

\begin{figure}[h]
  \centering
  \includegraphics[width=1\columnwidth]{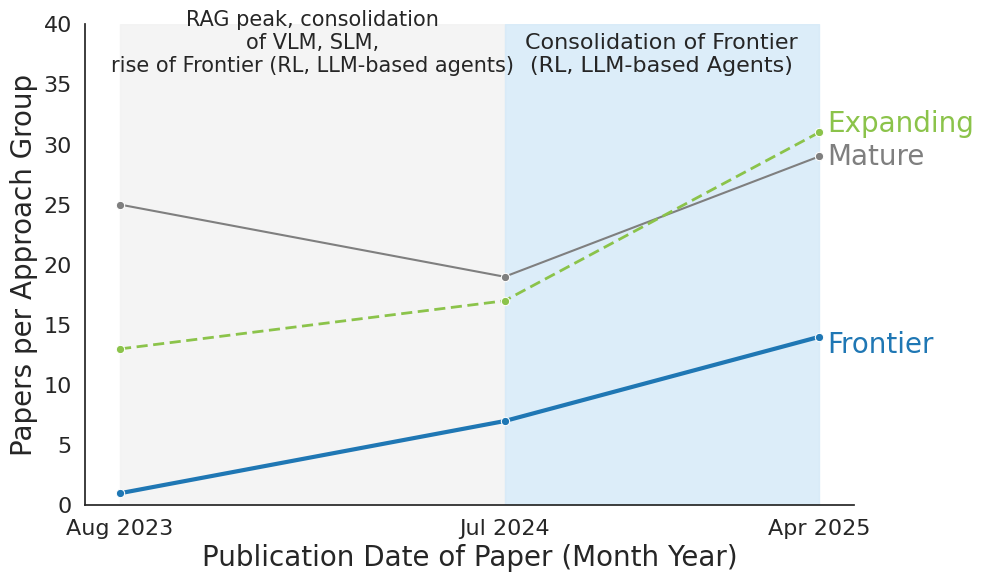}
  \caption{Latest trends in Financial NLP.}
  \label{fig:latest_trends}
\end{figure}

Financial NLP is entering a new phase, driven not only by LLMs but by a deeper understanding of their strengths and limitations. As techniques such as RAG and open-source fine-tuning become standard (grey line in Figure~\ref{fig:latest_trends}), multimodal models and small language models (green line) are gaining traction.
New trends are also emerging (blue line in Figure~\ref{fig:latest_trends}), most notably multi-agent systems, which range from simple expert--critic setups to more complex architectures.
At the same time, the gap between academic research and real-world financial practice remains an open question, as the field shifts its focus from question answering toward reasoning-oriented systems.

\section{Conclusion}

We presented a structured, quantitative meta-analysis of GenAI-driven transformation in Financial NLP, based on a corpus of 681 papers from 2022--2025. Our analysis reveals systematic shifts in tasks, datasets, risks, and architectural paradigms, marking a transition from model-centric experimentation to system-oriented design.

MetaGraph serves as the enabling framework that operationalizes ontology-guided LLM extraction at scale, producing a reproducible and queryable representation of the field. Beyond Financial NLP, this approach illustrates how structured knowledge graph construction can support data-driven monitoring of rapidly evolving research domains.

\section{Limitations}

\begin{itemize}
    \item Our approach relies on a \textit{manually defined ontology}, which introduces an inductive bias in how entities and relations are categorized. While this provides structure and interpretability, it may also limit flexibility and overlook alternative or emergent conceptualizations.
    \item Despite continuous human validation and refinement, the \textit{entity extraction and taxonomy induction processes remain based on LLMs}, which are inherently susceptible to hallucinations, inaccuracies, and bias. These limitations may affect both the precision and completeness of the extracted knowledge.
    \item The initial selection of papers was based on heuristic keyword search. This approach may have missed some papers that should be considered part of Financial NLP.
\end{itemize}

\section*{Acknowledgments}
\noindent We thank David Rosenberg for his detailed feedback, which substantially improved this paper.

\bibliography{custom}

\appendix

\section{The Geography of Financial NLP}
\label{apx:geography}

\begin{figure*}
    \centering
    \includegraphics[width=\textwidth]{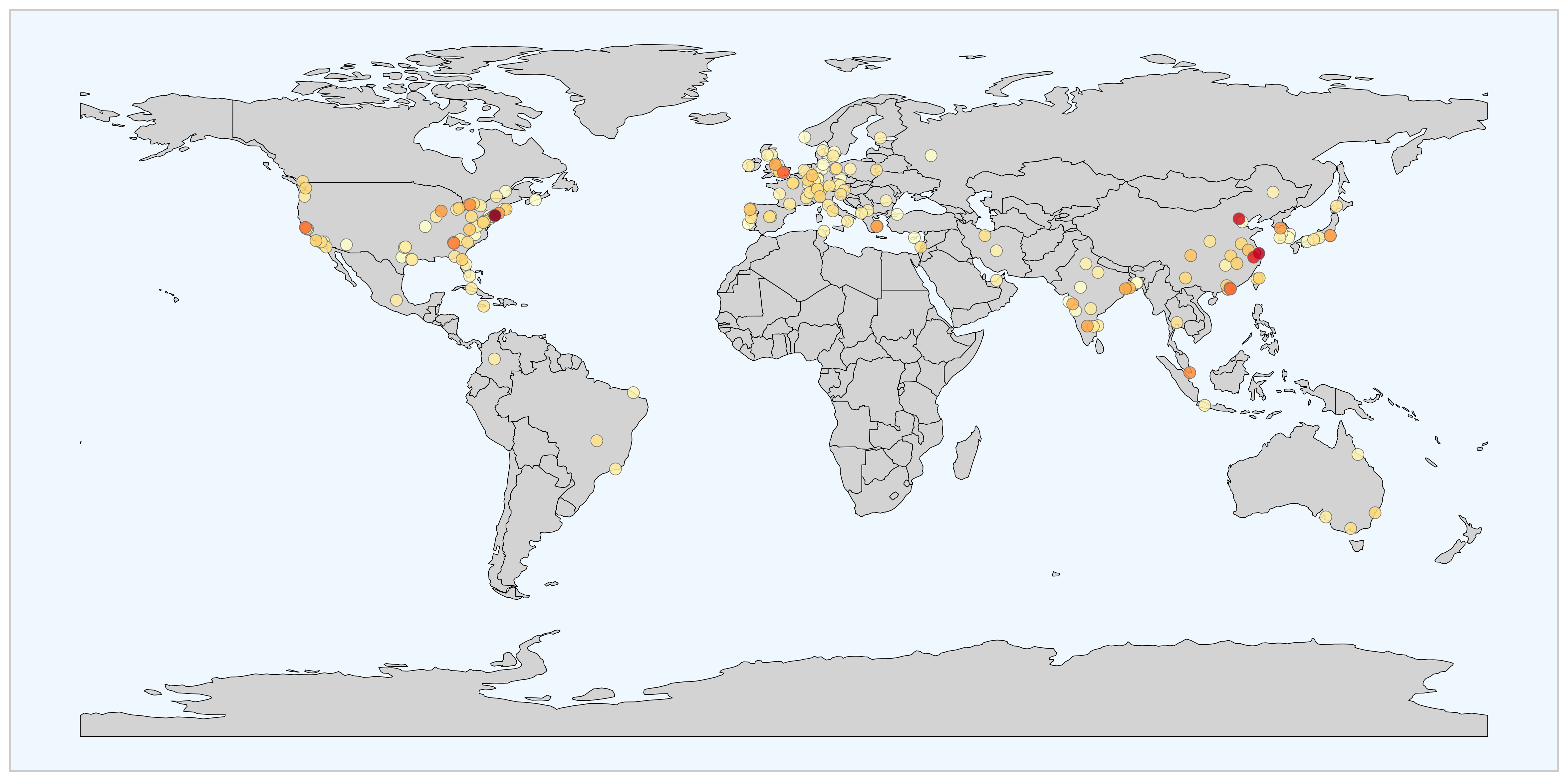}
    \caption{Geography of Financial NLP. The intensity of colors indicates the frequency of contributions involving an institution based in that place.}
    \label{fig:geography_financial_nlp}
\end{figure*}

The map depicted in Figure~\ref{fig:geography_financial_nlp} illustrates the geographical distribution of institutions represented in our corpus. It highlights that Financial NLP research predominantly clusters around three major global hubs. In the United States, research activity is highly concentrated along the Atlantic Coast, with a distinct epicenter in New York City. In East Asia, significant research centers have emerged in major economic and technological hubs, notably within China, Korea, Japan, Hong Kong, and Singapore. Europe, on the other hand, presents a different pattern: research activities are more dispersed, reflecting a fragmented landscape with multiple smaller centers rather than a single dominant hub. This contrasting geography suggests regional differences in collaboration patterns, institutional density, and possibly cultural or economic factors influencing research organization in Financial NLP.

\subsection{Financial NLP Graph Sub-trees}
\begin{figure*}
    \centering
    \includegraphics[width=\textwidth]{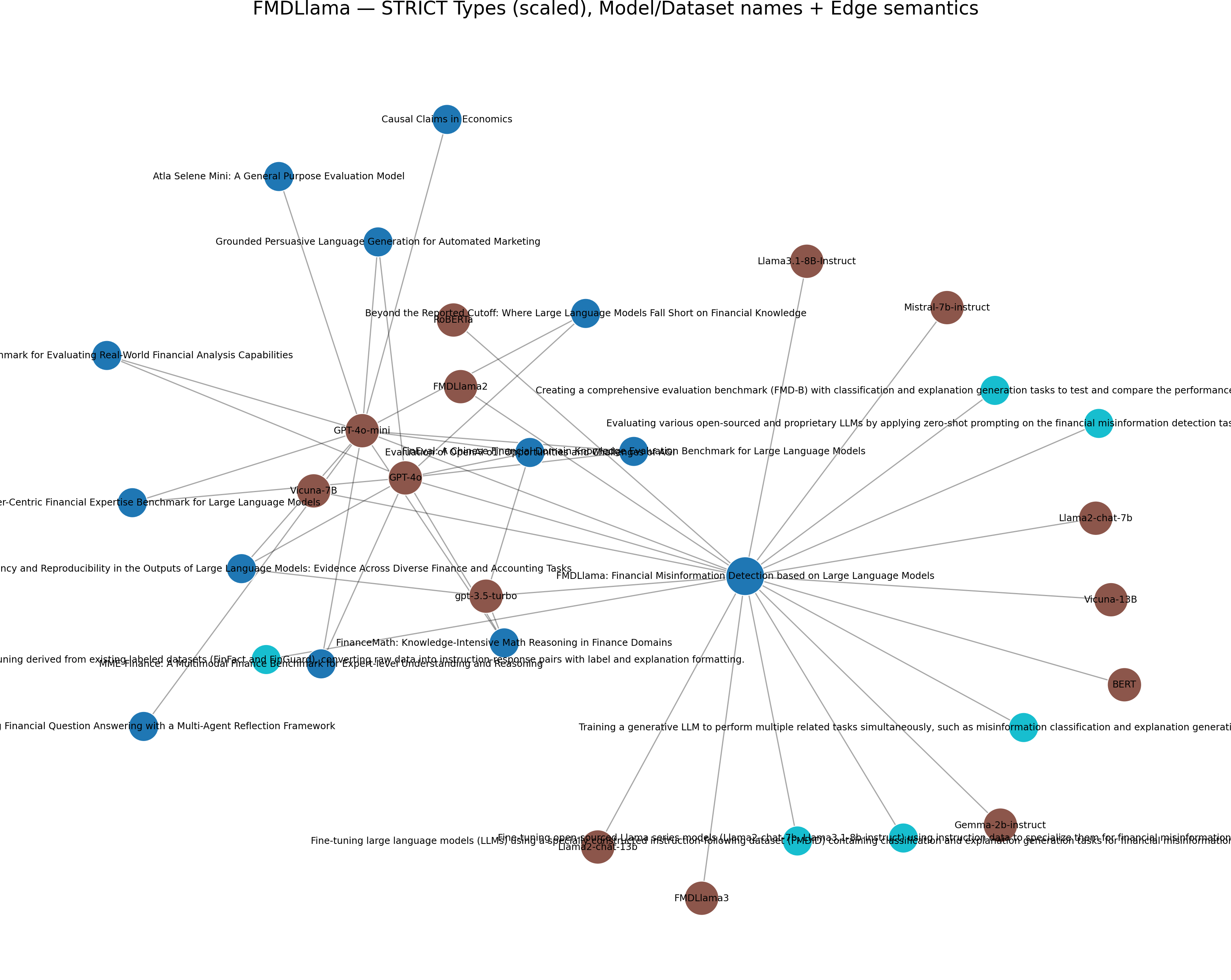}
    \caption{Subtree snippet of the Financial NLP graph structure with selected paper titles, models and dataset descriptions.}
    \label{fig:graph_subtree}
\end{figure*}

We show a snippet of the ontology emphasizing the connections between the papers on financial topics, the models used and datasets used. For readability, we do not include author names and institutions, paper motivations or limitations. Figure ~\ref{fig:graph_subtree} shows the subtree.

\section{Prompts for Graph Extraction and Enrichment}
\label{apx:prompts}
This section details the prompts designed for extracting and enriching the graph structure from the corpus of Financial NLP research papers. The prompts were crafted to ensure clarity, conciseness, and consistency in extracting different entity types and relationships, facilitating accurate and systematic analysis.

\subsection{Graph Extraction}
For entities such as limitations, motivations, and techniques, a unique standardized prompt format was adopted to maintain consistency across extraction tasks. Given an entity type represented by \textit{X} (e.g., limitations), the prompt follows the format shown in Listing~\ref{lst:prompt_lim_mot_tech}.

\begin{lstlisting}[float=*, floatplacement=h!, language={}, caption={Format of the prompt used for extraction of limitations, motivations, and techniques}, label={lst:prompt_lim_mot_tech}]
You will be given the full text of a paper that applies NLP techniques to financial data analysis. Your goal is to identify the {ENTITY TYPE} mentioned in the paper.
You should summarize each {ENTITY TYPE} in a brief and concise description (no more than 50) words.
\end{lstlisting}

A separate prompt, detailed in Listing~\ref{lst:prompt_task_data_model}, was developed to specifically extract tasks, datasets, and models. This distinct prompt was necessary to accurately capture relationships among these entities. Other types of relationships, such as entities co-occurring within the same paper, were implicitly inferred from extracted information and thus did not require additional prompting.

\begin{lstlisting}[float=*,language={}, caption={Format of the prompt used for extraction of tasks, datasets, models, and their relationships}, label={lst:prompt_task_data_model}]
You will be given the full text of an article that applies NLP to financial data. Your goal is to answer with the following information:

- General Reasoning: reason step-by-step to answer the following questions: which are the tasks the paper evaluates the models on? Which are the datasets used for each task?
Use the result of the previous reasoning to output:
- Tasks: a list of tasks the paper evaluates the models on. For each task, you have to specify:
    - Datasets: a list of datasets used to test the models on the task.
    For each dataset, you have to provide:
        - Dataset Reasoning: reasoning step-by-step to extract the following information
        - Dataset Name: the name of the dataset. If the dataset is referred generically, specify "generic"
        - Dataset Created: "yes" if the dataset is created by the authors, "no" otherwise
        - Reference name: name of the paper that proposes the dataset, if the reference paper is cited by the authors
        - Reference Year: Year of the paper that proposes the dataset, if the reference paper is cited by the authors where the dataset was extracted.
        - Signals: List of signals included in the dataset. Can be one or more from [Text, Tables, Image, Audio, Code, Time Series, Video, Charts, Equations]
        - Sources: List of sources from which the dataset is taken. Can be one or more from [News, Social Media/Forums, Company Reports, Company Fundamentals \& Indicators, Earnings Calls, Analyst Reports, University Textbooks, Financial Analyst Exams]
        - Annotation: The way the dataset was annotated. Can be one or more from [Manual, Synthetic, Spontaneous, Human-in-the-loop]
    - Models: a list of models that are evaluated on the task:
        - Models reasoning: reasoning step-by-step to extract the following information
        - Model Position: "main" if it is the model on which the main contribution of the authors is based, "comparison" if the model was used as a comparison.
        - Model Name: the name of the model
        - Reference Name: Name of the paper that proposes the model, if the reference paper is cited by the authors where the model was extracted.
        - Reference Year: Year of the paper that proposes the model, if the reference paper is cited by the authors where the model was extracted.
        - Size: Can be one of small (up to 8B parameters), medium (between 8B and 80B parameters), and large (more than 80B parameters).
        - Parameter Size: Number of parameters (in billions) if it is specified by the authors, None otherwise.
\end{lstlisting}

\subsection{Taxonomy Extraction}

To systematically derive taxonomies for specific entity types, we employed an iterative process, where multiple representative samples of entities were provided to the LLM using the prompt shown in Listing~\ref{lst:prompt_taxonomy_extraction}. This iterative approach allowed for the identification of recurrent patterns and commonly occurring entity types, ensuring robust and meaningful taxonomic categories.

\begin{lstlisting}[float=*, language={}, caption={Format of the prompt used for taxonomy extraction}, label={lst:prompt_taxonomy_extraction}]
You will be provided with a list of descriptions. Each description corresponds to a {ENTITY TYPE} extracted from a Financial NLP paper.
Your goal is to identify recurring patterns in the list of {ENTITY TYPE}. In particular, you should identify types of {ENTITY TYPE} that appear multiple times in the list. Do not try to find types of {ENTITY TYPE} that cover every single example-only identify those that recur with a certain frequency.
\end{lstlisting}

\subsection{Entity Classification}
To classify entities consistently across different categories, we utilized a universal prompt structure, as illustrated in Listing~\ref{lst:prompt_classification}. This approach ensures uniformity and accuracy in the classification of entities based on predefined categories.

\begin{lstlisting}[float=*, language={}, caption={Format of the prompt used for entity classification}, label={lst:prompt_classification}]
You will be given a description of an entity extracted from a Financial NLP papers. The entity is of type {ENTITY TYPE}. Your goal is to classify the entity into one or more of the following categories:

{LIST OF POSSIBLE CATEGORIES}
\end{lstlisting}

\section{Additional Material for Analysis}
\label{apx:analysis}
\begin{lstlisting}[float=*, language={}, caption={Prompt used for detection of the stance towards LLMs from motivations}, label={lst:prompt_llm_stance}]
You will receive a JSON array containing every *motivation paragraph* for ONE
research paper in NLP-for-Finance.

**Task (for EACH element, in the same order)**
1. Think step-by-step: does the paragraph primarily
    - show *enthusiasm / experimentation* with LLMs, **or**
    - propose *hybrid / alternative* solutions (i.e. highlight LLM limitations)?
    - If no stance towards LLMs is expressed, set Enthusiasm = "other"
2. Summarise your reasoning in 2-3 sentences.
\end{lstlisting}

This section provides supplementary analyses and detailed data supporting the insights discussed in the main body of the paper. It encompasses comprehensive overviews of task trends, evolving methodologies, and changing attitudes within Financial NLP research, presented through tables and figures to offer clear visual representations of temporal developments.

\subsection{Temporal Evolution of Financial NLP Tasks}
Table~\ref{tab:evolution_task_variants} presents a detailed percentage distribution of Financial NLP tasks across three different timeframes: January 2022--August 2023 (T1), September 2023--July 2024 (T2), and August 2024--April 2025 (T3). This analysis is based on categories derived by prompting Gemini-2.5-Flash to output a taxonomy from samples of task descriptions extracted from the paper, and we used the same model to assign each task to one of the categories.

The data illustrates limited innovation in information extraction: the tasks remain the same since the first period, and the percentage of papers practicing them is decreasing (with the exception of claim verification, which is a recent novelty). We can observe that the sharpest decline concerns financial sentiment analysis, which was the most practiced task in the first period and has experienced a significant drop in the last, reaching a percentage that is, by itself, lower than retrieval-enhanced QA. Finally, we can see that the true core of innovation is concentrated around financial QA. Tasks that were already practiced in the first period have attracted increasing attention over time, and new, more complex forms of QA (such \textbf{as} financial consulting and multimodal QA) have become widespread in the latest period.

\begin{table}[h]
\centering
\small
\setlength{\tabcolsep}{3pt}
\begin{tabularx}{\columnwidth}{
  >{\raggedright\arraybackslash}X
  S[table-format=2.2]
  S[table-format=2.2]
  S[table-format=2.2]
}
\toprule
\textbf{Task Category \& Name} & {\textbf{T1(\%)}} & {\textbf{T2(\%)}} & {\textbf{T3(\%)}} \\
\midrule
\multicolumn{4}{l}{\textbf{Information Extraction}} \\
\cmidrule(r){1-4}
Intent Detection & 1.99 & 1.73 & 1.73 \\
Stance Detection & 1.00 & 1.56 & 0.58 \\
NER & 7.21 & 8.67 & 5.32 \\
Relation Extraction & 7.21 & 4.16 & 4.03 \\
Semantic Annotation & 12.69 & 7.11 & 4.17 \\
Claim Verification & 1.24 & 1.21 & 3.31 \\
Event Based Text Annotation & 6.47 & 9.53 & 6.47 \\
Event Extraction & 3.23 & 2.08 & 0.86 \\
Argument Mining & 0.75 & 1.56 & 0.58 \\
\midrule
\multicolumn{4}{l}{\textbf{Financial Sentiment Analysis}} \\
\cmidrule(r){1-4}
Financial Sentiment Analysis & 15.67 & 15.77 & 9.78 \\
Financial Emotion Analysis & 0.50 & 0.87 & 0.43 \\
\midrule
\multicolumn{4}{l}{\textbf{Stock Market Prediction}} \\
\cmidrule(r){1-4}
Stock Price Change Prediction & 12.19 & 7.63 & 8.20 \\
Stock Volume Prediction & 0.25 & 0.35 & 0.58 \\
Stock Return Prediction & 4.73 & 3.47 & 2.59 \\
\midrule
\multicolumn{4}{l}{\textbf{QA \& Specialized Tasks}} \\
\cmidrule(r){1-4}
Retrieval Enhanced QA & 5.47 & 7.80 & 12.81 \\
Numerical QA & 5.47 & 7.45 & 9.50 \\
Conversational QA & 2.24 & 3.12 & 3.60 \\
Long Document QA & 3.23 & 4.51 & 7.34 \\
Causal QA & 1.99 & 1.04 & 2.30 \\
Tabular QA & 2.24 & 3.99 & 4.46 \\
Financial Consulting QA & 2.24 & 4.16 & 6.04 \\
Temporal Reasoning QA & 0.50 & 0.52 & 0.86 \\
Financial Terminology Explanation & 0.75 & 1.21 & 2.73 \\
Multimodal QA With Images & 0.75 & 0.52 & 1.73 \\
\bottomrule
\end{tabularx}
\caption{Percentage distribution of Financial NLP tasks across different timeframes (T1: Jan 2022--Aug 2023, T2: Sep 2023--Jul 2024, T3: Aug 2024--Apr 2025).}
\label{tab:evolution_task_variants}
\end{table}

\subsection{Trends in LLM Risks and Limitations}
\label{app:risks_and_limitations}
Table~\ref{tab:evolution_safety_concerns} categorizes the evolving risks associated with large language models (LLMs) in financial applications, as identified in the reviewed corpus. Recall that these categories are obtained by prompting Gemini-2.5-Flash to output a taxonomy from samples of safety limitations extracted from the paper, and we used the same model to assign the limitations to one of the categories.

Notably, safety concerns have sharply increased from the first to the second period, and \textbf{are} still sharply rising today. However, we can distinguish groups with different trends: concerns that were already prevalent, such as inaccuracy and the inability to adapt to new data, have become even more pronounced. At the same time, limitations rarely highlighted by paper authors before have emerged. These include the impact that well-known limitations of LLMs, such as bias, can have in the financial sector, as well as privacy and misuse issues.

\subsection{Evolution of Retrieval-Augmented Generation (RAG) Approaches}
Table~\ref{tab:rag_evolution} summarizes the distribution of different retrieval-augmented generation (RAG) configurations over the studied timeframes. It should be noted that we are using the distinguishing factors for RAG approaches as proposed in the survey by \citet{Gao2023RetrievalAugmentedGF}.

We can observe a clear trend that follows the expansion of data sources within the sector: the documents from which retrieval occurs are becoming increasingly longer, and the landscape of modalities is growing more diverse. In particular, there has been an increased use of structured data sources such as tables and databases. As RAG tasks become more complex and refined, the model is asked to retrieve more and more information, which is then incorporated into the prompt. This testifies to the progressive transformation of information extraction tasks into retrieval tasks.

\subsection{Shifts in Data Reuse Practices}
Figure~\ref{fig:created_data_evolution_2} illustrates the changing proportions of datasets explicitly referencing prior literature over time, relative to all datasets used in published studies. The trend suggests a shift in data reuse practices: following an initial phase characterized by widespread creation of new datasets, researchers increasingly began to adopt and build upon existing resources.

\subsection{Prompt Engineering Techniques}
Table~\ref{tab:evolution_prompt_engineering} highlights the evolving landscape of prompt engineering techniques over three distinct timeframes. The categories are taken from \cite{schulhoff2025promptreportsystematicsurvey}. 
We can see an explosion in the use of prompt engineering techniques in the period following the advent of LLMs. While zero-shot remains the most widespread and dominant technique, few-shot has seen a recent decline in usage, being increasingly replaced by newer techniques to improve output accuracy. Chain-of-thought prompting enhances the model's reasoning, improving performance on complex tasks and supporting interpretability. RAG adds contextual information to reduce inaccuracies and provide updated knowledge, while techniques such as self-criticism use another or the same LLM to check the output. These techniques share a common goal: they add complexity and sophistication to the prompting process to mitigate model shortcomings.

\subsection{Stance Detection towards LLMs in Papers Motivations}
\label{sec:Stance_detection}
In order to obtain a quantitative insight into how the stance towards LLMs has changed over time, we applied Gemini 2.5 Flash zero-shot on the papers' motivation, using the prompt in Listing~\ref{lst:prompt_llm_stance}.

\begin{figure}[h]
    \centering
    \includegraphics[width=1\columnwidth]{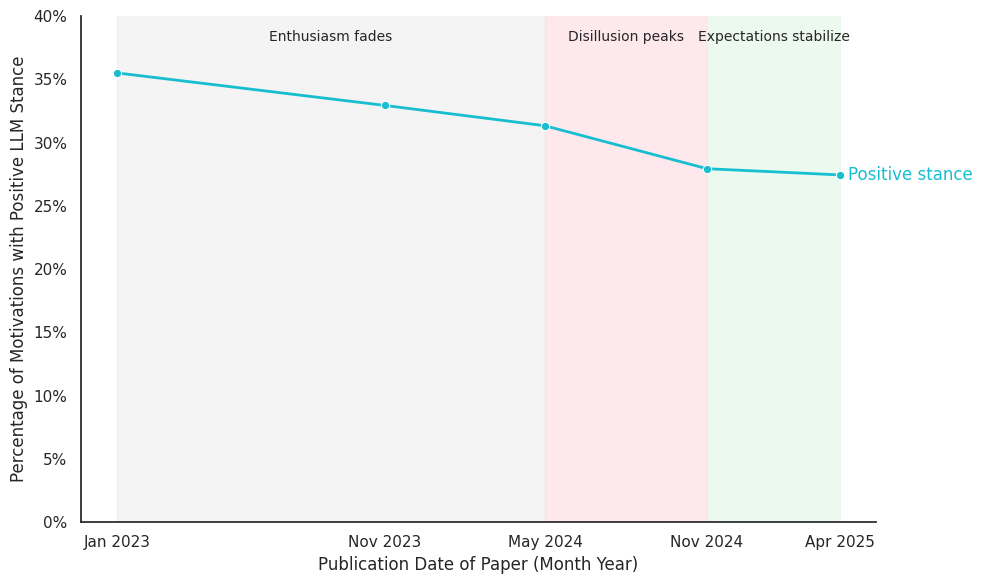}
    \caption{Attitude toward LLMs \textbf{has} evolved over time.}
    \label{fig:llm_stance_evolution}
\end{figure}

Figure~\ref{fig:llm_stance_evolution} provides a quantitative depiction of evolving attitudes toward LLMs over time, as reflected in the motivations stated in their work. We observe that papers framing LLMs as a solution to the limitations of earlier models have steadily declined, giving way to a growing number of studies motivated by the goal of mitigating the limitations of the LLMs themselves.



\end{document}